\documentclass[10pt,twocolumn,letterpaper]{article}

\usepackage{iccv}              

\definecolor{iccvblue}{rgb}{0.21,0.49,0.74}
\usepackage[pagebackref,breaklinks,colorlinks,allcolors=iccvblue]{hyperref}
\usepackage{makecell}
\usepackage{multirow}
\usepackage[accsupp]{axessibility}

\newcommand{\draftcomment}[2]{{\color{black}#2}}
\newcommand{\jk}[1]{\draftcomment{cyan}{#1}}
\newcommand{\zp}[1]{\draftcomment{red}{#1}}

\def\paperID{2740} 
\def\confName{ICCV}
\def\confYear{2025}

\title{Rethinking Discrete Tokens: Treating Them as Conditions for \\ Continuous Autoregressive Image Synthesis}

\author{
Peng Zheng$^{1,2}$ \quad
Junke Wang$^3$ \quad
Yi Chang$^1$ \quad
Yizhou Yu$^4$ \quad
Rui Ma$^{1,}$\thanks{Corresponding authors} \quad
Zuxuan Wu$^{2,3,}$\footnotemark[1] \\
$^1$School of Artificial Intelligence, Jilin University \quad
$^2$Shanghai Innovation Institute \\
$^3$Institute of Trustworthy Embodied AI, Fudan University \\
$^4$Department of Computer Science, The University of Hong Kong \\
}

\begin{document}
\maketitle

\begin{abstract}
Recent advances in large language models (LLMs) have spurred  interests in encoding images as discrete tokens and leveraging autoregressive (AR) frameworks for visual generation. However, the quantization process in AR-based visual generation models inherently introduces information loss that degrades image fidelity. To mitigate this limitation, recent studies have explored to autoregressively predict continuous tokens. Unlike discrete tokens that reside in a structured and bounded space, continuous representations exist in an unbounded, high-dimensional space, making density estimation more challenging and increasing the risk of generating out-of-distribution artifacts. Based on the above findings, this work introduces \textbf{DisCon} (Discrete-Conditioned Continuous Autoregressive Model), a novel framework that reinterprets discrete tokens as conditional signals rather than generation targets. By modeling the conditional probability of continuous representations conditioned on discrete tokens, DisCon circumvents the optimization challenges of continuous token modeling while avoiding the information loss caused by quantization. DisCon achieves a gFID score of \textbf{1.38} on ImageNet 256$\times$256 generation, outperforming state-of-the-art autoregressive approaches by a clear margin. Project page: \url{https://pengzheng0707.github.io/DisCon}.

\end{abstract}


\section{Introduction}
\label{sec:intro}


\begin{figure}
\centering
\includegraphics[width=0.9\linewidth]{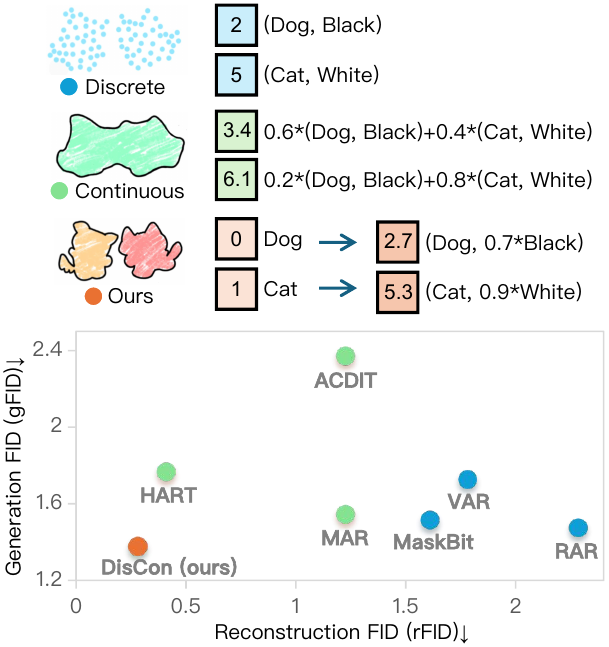}
\caption{\textbf{Visual Data Representations.} Discrete AR models represent data as separate categories, simplifying learning but introducing quantization-induced information loss, leading to higher rFID. In contrast, continuous AR models assume data lies in a continuous space, achieving lower rFID. However, unlike discrete tokens that reside in a structured and bounded space, continuous representations exist in an unbounded, high-dimensional space, increasing the risk of generating out-of-distribution artifacts, which limits improvements in gFID. Our approach models data as a finite set of disjoint continuous representations, using discrete tokens to determine the broader structure and continuous tokens to refine details, effectively reducing optimization difficulty while achieving both low rFID and gFID.}
\label{fig:hybrid}
\end{figure}

\begin{figure*}
    \centering
    \includegraphics[width=\linewidth]{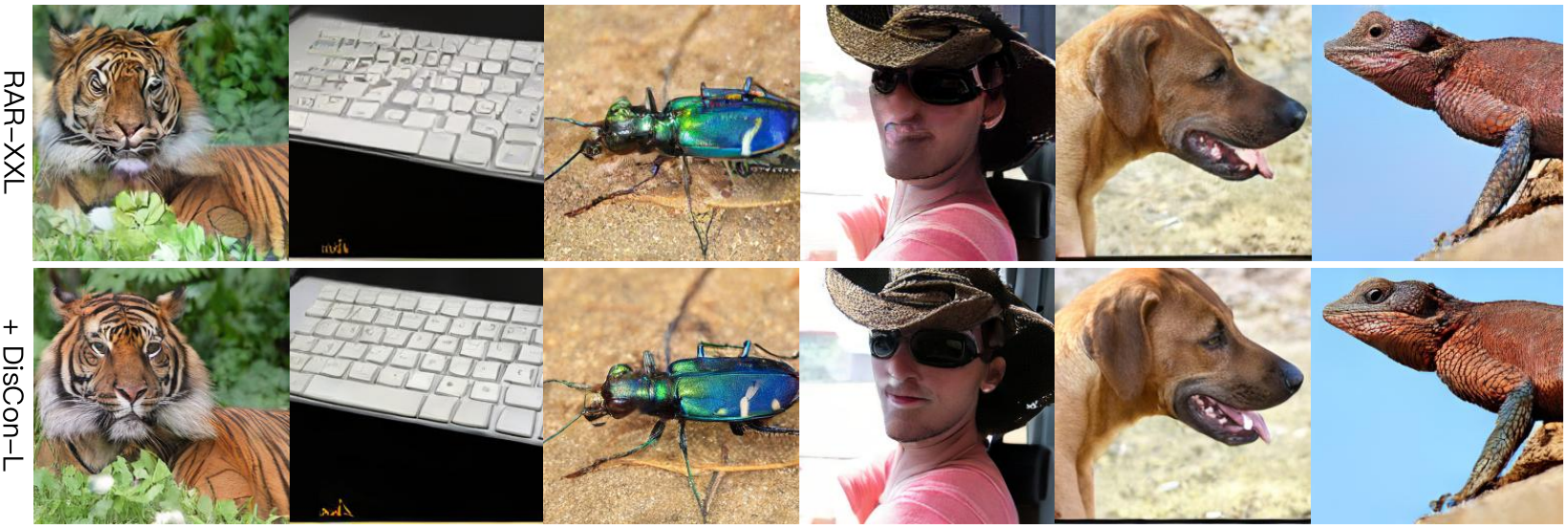}
    \caption{\textbf{Discrete vs. Continuous AR Models. } Top: Images generated by RAR-XXL~\cite{yu2024randomized}, the SOTA discrete AR model. Bottom: Images generated by our DisCon-L model, a continuous AR model conditioned on the discrete tokens produced by RAR-XXL. Zoom in for better visualization to observe the significant improvements in generation quality.}
    \label{fig:rar}
\end{figure*}


Image generation has long been a central topic in artificial intelligence. More recently, the remarkable success of large language models (LLMs)~\cite{ouyang2022training,touvron2023llama,team2023gemini,achiam2023gpt,liu2024deepseek} has reignited interest in \emph{autoregressive} (AR)-based image generation~\cite{yu2021vector}, offering a promising path towards general multimodal LLMs~\cite{zhou2024transfusion,shi2024llamafusion,tong2024metamorph,wu2024vila,xie2024show,wang2024emu3,team2024chameleon,qu2024tokenflow,wu2024janus}. These methods first quantized images into discrete tokens, and then apply autoregressive transformers to predict them in a sequential manner. However, the quantization step inevitably discards some visual information, thereby constraining the fidelity of the generated images.

To address this issue, continuous autoregressive models have been explored to avoid the information loss of discrete tokenization~\cite{li2025autoregressive,hu2024acdit,tang2024hart,ren2024flowar}. These models directly learn continuous latent representations but face optimization difficulties, as the space of continuous tokens is inherently more complex than that of discrete tokens. Consequently, current continuous AR methods often lag behind their discrete counterparts in generation performance, despite theoretically richer representations.

Our key observation is that real-world image datasets can be viewed as \emph{finite collection of disjoint continuous distributions}. Each local mode of the data distribution corresponds to a distinct region in the latent space, separated from other modes by clear gaps. For example, images of different object categories (\eg, different animal species) form well-separated modes, while attributes such as color or texture vary smoothly within each mode. This perspective suggests that high-level category-like structure can be modeled by discrete tokens, whereas fine-grained variability is better captured by continuous tokens. Figure~\ref{fig:hybrid} provides a schematic illustration of this data representation compared to existing methods.

Inspired by this, we propose \textbf{DisCon}, a novel autoregressive framework that reinterprets discrete tokens as high-level conditions for continuous generation. In our approach, the modeling task is decoupled into two steps: (i) predicting discrete tokens to identify the local mode of the data to synthesize and (ii) predicting continuous tokens to refine fine-grained details within that mode, conditioned on the discrete tokens. Since discrete tokens can be reliably learned and already capture most essential information, the subsequent continuous modeling becomes significantly simpler. This design leverages the powerful representational capacity of continuous distributions while reducing optimization difficulty, resulting in improved generation quality across both fidelity and reconstruction metrics.

In summary, our contributions are as follows:
\begin{itemize}
\item We propose a novel discrete-as-condition paradigm which treats discrete tokens not as generation targets but as structural priors that steer a continuous AR model. This perspective naturally handles data that can be described as a finite set of disjoint continuous distributions.
\zp{\item We introduce DisCon, a novel two-stage pipeline that first models a discrete distribution and then estimates the conditional probability distribution from discrete to continuous representations. This design avoids the difficulty of directly modeling continuous distributions, enabling high-quality image generation.}
\item Experiments demonstrate that DisCon achieves superior performance on generation fidelity (1.38 gFID) and reconstruction accuracy (0.28 rFID) on ImageNet-256~\cite{deng2009imagenet}, outperforming leading AR baselines while maintaining fast inference speeds.
\end{itemize}


\section{Related Work}
\label{sec:related}


\subsection{Overview of Image Generation Paradigms} Early image synthesis was dominated by Generative Adversarial Networks (GANs)~\cite{goodfellow2014generative} and Variational Autoencoders (VAEs)~\cite{kingma2013auto}, which directly map noise to data distributions. More recently, diffusion models~\cite{rombach2022high,podell2023sdxl,dhariwal2021diffusion,ma2025inference,gao2023mdtv2} have demonstrated impressive results by iteratively denoising random inputs to generate high-quality images. However, autoregressive (AR) approaches have gained renewed attention due to their compatibility with large language models (LLMs)~\cite{ouyang2022training,touvron2023llama,team2023gemini,achiam2023gpt,liu2024deepseek}, offering a unified framework for multimodal generation.

\subsection{Discrete Autoregressive Image Generation}
Inspired by the success of large language models~\cite{ouyang2022training,touvron2023llama,team2023gemini,achiam2023gpt,liu2024deepseek} in modeling discrete sequences, autoregressive (AR) approaches have been adapted to image generation~\cite{yu2021vector,lee2022autoregressive,pang2024randar,han2024infinity,gu2024rethinking,zhu2024scaling,yu2025image,wang2024parallelized,shi2024taming,yu2023language,sun2024autoregressive}. These methods quantize images into sequences of discrete tokens (\eg , via VQGAN~\cite{yu2021vector}) and model the token distribution sequentially using cross-entropy loss. While this enables efficient training and fast inference, the quantization process inevitably discards fine-grained image details, limiting reconstruction fidelity. Notable examples include RAR~\cite{yu2024randomized}, which refines generation by permuting token sequences for richer context, and VAR~\cite{tian2025visual}, which models image structure through next-scale prediction. However, the inherent limitations of discrete tokens restrict the expressive capacity needed for high-fidelity synthesis.

\subsection{Continuous Autoregressive Image Generation}
To mitigate the drawbacks of quantization, recent research has explored generating continuous latent representations directly~\cite{chen2024softvq,yuan2024car,fan2024fluid}. For example, MAR~\cite{li2025autoregressive} incorporates a diffusion loss to model continuous latent variables, thereby enhancing representational fidelity. More recently, methods like FlowAR~\cite{ren2024flowar} combine AR modeling with flow matching techniques~\cite{lipman2022flow,dao2023flow} to generate continuous latent representations from a VAE, and ACDIT~\cite{hu2024acdit} fuses diffusion processes with AR to refine latent representations. Despite these advances, optimizing continuous AR models remains challenging due to the inherent complexity of continuous spaces. Additionally, HART~\cite{tang2024hart} proposes to learn both discrete and continuous representations within a single model; however, it still treats discrete tokens as generation targets, which is fundamentally different from our approach.


\subsection{Positioning Our Work}
Unlike previous approaches that treat discrete tokens as final outputs or directly model complex continuous spaces, our work rethinks their role in AR image generation. We propose to use discrete tokens purely as high-level conditional signals for a continuous AR model, effectively transforming the original complex modeling problem into two simpler tasks: (i) learning a discrete distribution and (ii) modeling the conditional distribution from discrete to continuous space. Since discrete tokens already capture most of the essential information, the remaining conditional modeling becomes significantly easier, leading to improved optimization stability and generation fidelity while preserving the strong expressive power of continuous representations.
\section{Method}
\label{sec:method}


\begin{figure*}[t]
\centering
\includegraphics[width=\linewidth]{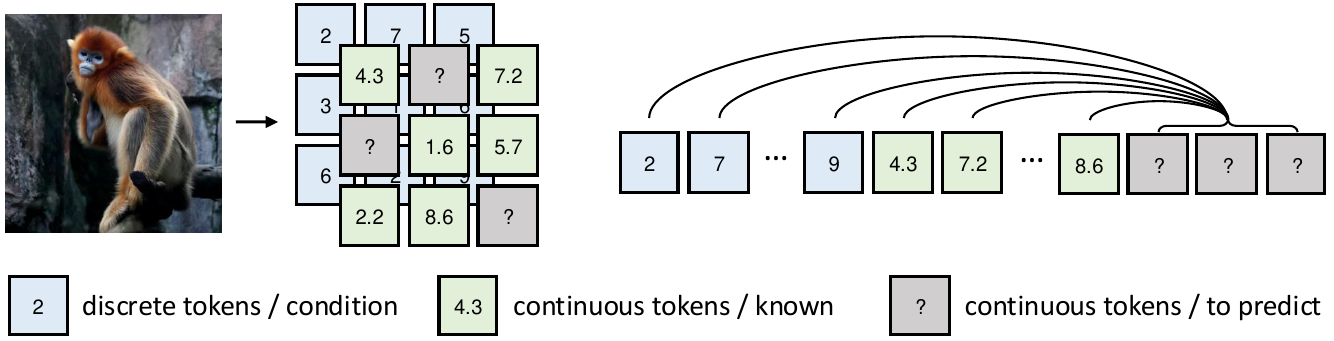}
\caption{\textbf{Overview of the Proposed DisCon Pipeline.} Given an input image, discrete and continuous tokens are first extracted using pre-trained tokenizers, with a certain proportion of the continuous tokens masked. An autoregressive model then predicts the masked tokens, conditioned on both the discrete tokens and the available continuous tokens. During inference, a pre-trained discrete AR model (\eg, RAR-XXL~\cite{yu2024randomized}) first generates the conditional discrete tokens, which guide the continuous AR model in producing high-fidelity continuous tokens that are finally decoded into the output image.}
\label{fig:pipeline}
\end{figure*}

The proposed framework, \textbf{DisCon} (Discrete-Conditioned Continuous Autoregressive Model), \jk{is a two-stage image generation framework that bridges the gap between discrete and continuous autoregressive image generation}. It decouples the prediction of global structure from fine-detail synthesis by employing discrete tokens solely as high-level conditional signals to guide a continuous AR model.


\subsection{Motivation \& Insight}
Natural images can be viewed as samples from a finite set of disjoint continuous distributions. For example, different object categories form distinct structural modes, while variations within each mode—such as color, texture, or shading—encode fine details. 
\jk{Inspired by this, this work decouples image generation into two stages: first, predicting the coarse global structure as discrete tokens, and then synthesizing fine-grained details using continuous tokens. This separation not only alleviates the optimization challenges associated with modeling high-dimensional continuous spaces but also circumvents the fidelity bottleneck imposed by direct quantization.}

\subsection{Preliminaries}
\paragraph{Discrete Autoregressive Models.}  
Discrete AR models factorize the joint probability over a sequence of tokens $\mathbf{x}_d = \{x_{d,1}, x_{d,2}, \ldots, x_{d,M}\}$ as:
\begin{equation}
    p(\mathbf{x}_d) = \prod_{i=1}^{M} p(x_{d,i} | x_{d,<i}),
\end{equation}
and are trained using the cross-entropy loss:
\begin{equation}
    \mathcal{L}_{\text{AR}} = -\sum_{i=1}^{M} \log p(x_{d,i} | x_{d,<i}).
\end{equation}
While methods such as RAR~\cite{yu2024randomized} and VAR~\cite{tian2025visual} effectively capture image structure, the quantization process inevitably discards fine details.

\paragraph{Masked Autoregressive Models.}  
\zp{Masked Autoregressive Models (MAR)~\cite{li2025autoregressive} propose applying autoregressive models in a continuous-valued space and adopting the masking strategy from Masked AutoEncoders (MAE)~\cite{he2022masked}, where the goal is to predict masked continuous tokens from the unmasked ones.}
However, directly predicting the continuous token representation $\mathbf{x}_c$ is challenging due to the high complexity of continuous distributions. To address this, MAR adopts a two-stage prediction strategy, where an autoregressive model is first employed to predict the intermediate latent variable $\mathbf{z}$ for the masked regions, and then a lightweight diffusion head uses $\mathbf{z}$ as a conditional signal to refine it into the final predictions $\mathbf{x}_c$. \jk{The learning of autoregressive transformer and the diffusion head is supervised by the diffusion loss}:
\begin{equation}
\label{eq:diffusion}
    \mathcal{L}(\mathbf{z},\mathbf{x}_c) = \mathbb{E}_{\varepsilon,t} \left[ \left\| \varepsilon - \varepsilon_\theta(\mathbf{x}_{c,t} | t, \mathbf{z}) \right\|^2 \right],
\end{equation}
where $\varepsilon$ is sampled from a standard normal distribution and $t$ denotes the noise schedule. This two-stage process effectively reduces the complexity of direct autoregressive prediction while preserving fine-grained image details. 
\zp{However, since MAR directly models continuous tokens, it remains challenging to learn a stable and accurate distribution due to the unbounded nature of continuous representations. This limitation motivates our approach to introduce discrete tokens as structured conditional signals.}

\subsection{Proposed Method: DisCon}
Our framework, \textbf{DisCon} (Discrete-Conditioned Continuous Autoregressive Synthesis), is designed to overcome the inherent difficulties of modeling continuous distributions. It extends MAR by using discrete tokens as high-level conditional signals, as shown in Figure~\ref{fig:pipeline}. Given an input image $\mathbf{I}$, we extract two representations via pre-trained tokenizers~\cite{yao2025reconstruction,chang2022maskgit}:

\begin{equation}
\mathbf{x}_d = \text{DiscreteTokenizer}(\mathbf{I}), \mathbf{x}_c = \text{ContinuousTokenizer}(\mathbf{I}).
\end{equation}
\zp{Here, $\mathbf{x}_d$ represents the extracted discrete tokens, while $\mathbf{x}_c$ consists of continuous tokens that retain richer representational capacity due to the absence of quantization. During training, we mask a portion of the continuous tokens and predict them using the complete discrete tokens and the unmasked continuous tokens. This allows the model to learn the conditional probability from discrete tokens to continuous tokens. Note that the prediction of continuous tokens is carried out through an intermediate latent variable $\mathbf{z}$. This latent variable helps to reduce the optimization complexity, as utilized in the MAR framework.}


Formally, the continuous AR model learns the conditional distribution: 
\begin{equation} 
p(\mathbf{x}_c | \mathbf{x}_d) = \prod_{i=1}^{M} p(z_i | \mathbf{x}_d,\mathbf{x}_{c,<i}) \cdot p(x_{c,i} | z_i), 
\end{equation} 
where $\mathbf{x}_c = {x_{c,1}, x_{c,2}, \ldots, x_{c,M}}$ represents the continuous tokens. The first term, $p(z_i | \mathbf{x}_d,\mathbf{x}_{c,<i})$, models the prediction of latent variables $z_i$ conditioned on discrete tokens $\mathbf{x}_d$ and available continuous tokens $\mathbf{x}_{c,<i}$, and the second term, $p(x_{c,i} | z_i)$, corresponds to the generation of the continuous token $x_{c,i}$ conditioned on the predicted latent variable $z_i$.

Here, $p(x_{c,i} | z_i)$ is modeled by a diffusion process, where the latent variable $z_i$ is fed into a diffusion head to generate the final continuous token $x_{c,i}$. This process enables the generation of fine-grained details from the latent space and is supervised by the loss defined in Equation~\ref{eq:diffusion}.

At inference, as ground-truth discrete tokens are unavailable, a pre-trained discrete AR model (\eg, RAR-XXL) generates an approximate token sequence:
\begin{equation}
\hat{\mathbf{x}}_d = \text{DiscreteAR}().
\end{equation}
These tokens condition the continuous AR model, which predicts the intermediate latent sequence $\mathbf{z}$. After diffusion refinement, the final image $\mathbf{I}^*$ is produced via a decoder:
\begin{equation}
\mathbf{x}_c = \text{ContinuousAR}(\hat{\mathbf{x}}_d),
\end{equation}
\begin{equation}
\mathbf{I}^* = \text{Decoder}(\mathbf{x}_c).
\end{equation}
\newline
\textbf{Why DisCon Works.}  
We decompose the modeling of continuous tokens into two subproblems. First, we model the discrete tokens $\mathbf{x}_d$ using a discrete AR model trained with a cross-entropy loss, a formulation that has been proven effective in prior work. Second, we model the conditional probability from discrete tokens to continuous tokens, i.e., $p(\mathbf{x}_c | \mathbf{x}_d)$, using a continuous AR model. Formally, we factorize the overall distribution as
\begin{equation}
    p(\mathbf{x}_c) = \sum_{\mathbf{x}_d} p(\mathbf{x}_c | \mathbf{x}_d) \, p(\mathbf{x}_d).
\end{equation}
Since the discrete tokens \( \mathbf{x}_d \) already capture most of the essential information in \( \mathbf{x}_c \) and can be effectively learned by the discrete AR model, the conditional probability \( p(\mathbf{x}_c | \mathbf{x}_d) \) becomes significantly simpler compared to directly modeling \( p(\mathbf{x}_c) \). Consequently, our method can effectively model continuous tokens, leading to improved generation performance.\\\\
\textbf{Architecture \& Flexibility.} DisCon features a modular architecture in which each component can be independently improved or replaced, including the tokenizers, the continuous AR transformer, and the pre-trained discrete AR model. This flexibility allows for the integration of more advanced transformer designs or stronger tokenizers, and makes DisCon readily adaptable to various image generation tasks or even other modalities.


\section{Experiments}
\label{sec:experiments}

\subsection{Implementation Details}
\paragraph{Dataset.}  
All models are trained on the ImageNet-256 dataset, which consists of 1,281,167 images. The dataset is augmented following the protocol in MAR~\cite{li2025autoregressive} by applying image flipping. The images are pre-tokenized using both a discrete tokenizer and a continuous tokenizer. The discrete tokenizer is adopted from MaskGIT~\cite{chang2022maskgit} (as used in RAR~\cite{yu2024randomized}), while for the continuous tokenizer we leverage the VA-VAE proposed in LightningDiT~\cite{yao2025reconstruction}.
\paragraph{Evaluation Setting.}
Quantitative metrics, including FID, IS, Precision, and Recall, are computed on 50k generated images. For our method, each image is conditioned on the discrete tokens generated by RAR-XXL. The discrete tokens are produced with the default classifier-free guidance (CFG) configuration in RAR, while the continuous tokens are generated without CFG. The default number of AR steps used in our method is 16. The results of other methods are taken from their respective papers.
\paragraph{Model Design.}  
The trainable continuous AR model is adopted from MAR-L. To facilitate adaptability to other continuous AR models, we modify the model only to incorporate conditioning on discrete tokens. To explore scalability, we propose two variants: DisCon-B and DisCon-L, with 427M and 558M parameters respectively.

\begin{table}[t]
    \centering
    \tabcolsep=0.03cm
    \begin{tabular}{l|cc|cccc}
    \toprule
         Method & rFID & Params & gFID$\downarrow$ & IS$\uparrow$ & Pre.$\uparrow$ & Rec.$\uparrow$ \\
         \toprule
         \multicolumn{7}{c}{\textcolor{gray}{Diffusion Models}} \\
         \textcolor{gray}{DiT~\cite{peebles2023scalable}} & \textcolor{gray}{0.61} & \textcolor{gray}{675M} & \textcolor{gray}{2.27} & \textcolor{gray}{278.2} & \textcolor{gray}{0.83} & \textcolor{gray}{0.57} \\
         \textcolor{gray}{SiT~\cite{ma2024sit}} & \textcolor{gray}{0.61} & \textcolor{gray}{675M} & \textcolor{gray}{2.06} & \textcolor{gray}{270.3} & \textcolor{gray}{0.82} & \textcolor{gray}{0.59} \\
         \textcolor{gray}{REPA~\cite{yu2024representation}} & \textcolor{gray}{0.61} & \textcolor{gray}{675M} & \textcolor{gray}{1.42} & \textcolor{gray}{305.7} & \textcolor{gray}{0.80} & \textcolor{gray}{0.64} \\
         \textcolor{gray}{LightningDiT~\cite{yao2025reconstruction}} & \textcolor{gray}{0.28} & \textcolor{gray}{675M} & \textcolor{gray}{1.35} & \textcolor{gray}{295.3} & \textcolor{gray}{0.79} & \textcolor{gray}{0.64} \\
         \textcolor{gray}{MaskDiT~\cite{zheng2023fast}} & \textcolor{gray}{0.61} & \textcolor{gray}{675M} & \textcolor{gray}{2.28} & \textcolor{gray}{276.6} & \textcolor{gray}{0.80} & \textcolor{gray}{0.61} \\
         \textcolor{gray}{MDTv2~\cite{gao2023mdtv2}} & \textcolor{gray}{0.61} & \textcolor{gray}{675M} & \textcolor{gray}{1.58} & \textcolor{gray}{314.7} & \textcolor{gray}{0.79} & \textcolor{gray}{0.65} \\
         \midrule
         \multicolumn{7}{c}{Discrete AR Models} \\
         VAR-d30-re~\cite{tian2025visual} & 1.78 & 2.0B & 1.73 & 350.2 & 0.82 & 0.60 \\
         RAR-B~\cite{yu2024randomized} & 2.28 & 261M & 1.95 & 290.5 & 0.82 & 0.58 \\
         RAR-L~\cite{yu2024randomized} & 2.28 & 461M & 1.70 & 299.5 & 0.81 & 0.60 \\
         RAR-XXL~\cite{yu2024randomized} & 2.28 & 1.5B & 1.48 & 326.0 & 0.80 & 0.63 \\
         MaskBit~\cite{weber2024maskbit} & 1.61 & 305M & 1.52 & 328.6 & - & - \\
         TiTok~\cite{yu2025image} & 1.71 & 287M & 1.97 & 281.8 & - & - \\
         RandAR-XXL~\cite{pang2024randar} & 2.19 & 1.4B & 2.15 & 321.97 & 0.79 & 0.62 \\
         MAGVIT-v2~\cite{yu2023language} & - & 307M & 1.78 & 319.4 & - & - \\
         LlamaGen-3B~\cite{sun2024autoregressive} & 0.94 & 3.1B & 2.18 & 263.3 & - & - \\
         \midrule
         \multicolumn{7}{c}{Continuous AR Models} \\
         FlowAR-L~\cite{ren2024flowar} & - & 589M & 1.90 & 281.4 & 0.83 & 0.57 \\
         FlowAR-H~\cite{ren2024flowar} & - & 1.9B & 1.65 & 296.5 & 0.83 & 0.60 \\
         MAR-B~\cite{li2025autoregressive} & 1.22 & 208M & 2.31 & 281.7 & 0.82 & 0.57 \\
         MAR-L~\cite{li2025autoregressive} & 1.22 & 479M & 1.78 & 296.0 & 0.81 & 0.60 \\
         MAR-H~\cite{li2025autoregressive} & 1.22 & 943M & 1.55 & 303.7 & 0.81 & 0.62 \\
         HART-d24~\cite{tang2024hart} & 0.41 & 1.0B & 2.00 & 331.5 & - & - \\
         HART-d30~\cite{tang2024hart} & 0.41 & 2.0B & 1.77 & 330.3 & - & - \\
         ACDIT-H~\cite{hu2024acdit} & 1.22 & 954M & 2.37 & 273.3 & 0.82 & 0.57 \\
         \midrule
         DisCon-B & 0.28 & 427M & 1.41 & 321.7 & 0.79 & 0.65  \\
         DisCon-L & 0.28 & 558M & \textbf{1.38} & 325.1  & 0.79  &  0.64 \\
    \bottomrule
    \end{tabular}
    \caption{Quantitative comparisons on ImageNet-256. Our method achieves SOTA performance among AR models. Among the metrics, gFID is the most important metric for evaluating the fidelity and diversity of the synthesis result. \emph{Params} denotes the number of trainable parameters. }
    \label{tab:comparisons}
\end{table}
\begin{figure*}
    \centering
    \includegraphics[width=\linewidth]{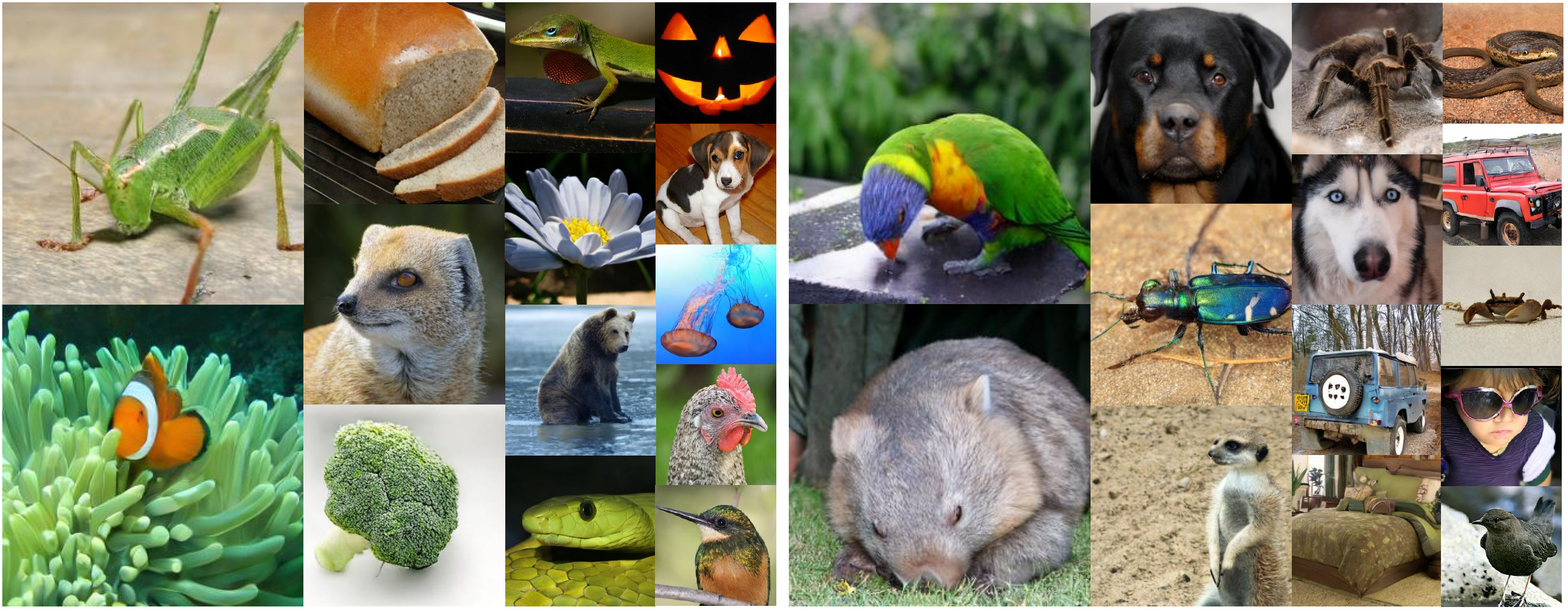}
    \caption{\textbf{Qualitative Results.} Images generated by DisCon-B (left) and DisCon-L (right), demonstrating high-fidelity synthesis.}
    \label{fig:disco}
\end{figure*}
\subsection{Main Results}
We compare DisCon against SOTA visual AR models on the ImageNet-256 dataset. As shown in Table~\ref{tab:comparisons}, continuous AR models employing stronger tokenizers achieve lower rFID values (\eg , 1.22 and 0.41). However, due to optimization complexity, their gFID values remain higher (\eg , 1.55 for MAR~\cite{li2025autoregressive}, which is the best among continuous models), showing a significant gap with rFID values. On the other hand, discrete AR models can even achieve lower gFID values (\eg , 1.48 for RAR~\cite{yu2024randomized}) than rFID values (\eg , 2.28 for RAR), since they are able to effectively model the discrete representations; nevertheless, they are limited by the representational power of discrete tokenizers. Instead, our proposed DisCon conditions continuous AR models on well-learned discrete tokens, achieving the best gFID value (1.38). We also provide qualitative comparisons in Figure~\ref{fig:rar}, where our continuous DisCon-L model demonstrates significant improvements over the discrete AR model RAR-XXL.

Among continuous AR models, our method builds on MAR by incorporating discrete token conditioning—a key enhancement that we further analyze in our ablation studies. Unlike HART~\cite{tang2024hart}, which learns both discrete tokens and residual continuous tokens by conditioning each on the corresponding discrete token yet still treats discrete tokens as generation targets, our approach treats discrete tokens solely as high-level conditions. This design simplifies the optimization process and leads to superior performance: our method achieves a gFID of 1.38, compared to HART's 1.77 gFID. Moreover, ACDIT~\cite{hu2024acdit}, which combines AR and diffusion models, and FlowAR~\cite{ren2024flowar}, which integrates AR with flow matching, both face similar optimization challenges as MAR, resulting in gFIDs of 2.37 and 1.65, respectively. Overall, by treating discrete tokens as conditions rather than as generation targets, our method attains better generation quality while preserving the strong representational power of continuous tokenizers.

Additional comparisons with diffusion models are also reported in Table~\ref{tab:comparisons}. Our method not only surpasses most leading diffusion models in quality metrics such as generation FID and reconstruction FID, but also achieves performance competitive with current SOTA methods. 
Notably, the sequential nature of AR models makes our approach inherently compatible with LLMs, paving the way for seamless integration into multimodal LLMs for joint vision-language tasks. In summary, the superior image quality and strong LLM compatibility underscore the robustness and versatility of our method.

We also explored model scaling in Tables~\ref{tab:comparisons} and \ref{tab:mar}. Notably, our DisCon-B with 427M parameters already achieves significant improvements over existing methods. Additional parameters, especially when integrated with LLMs, may lead to further performance gains. Finally, qualitative results generated by DisCon-B and DisCon-L are presented in Figure~\ref{fig:disco}.

    
    

\subsection{Ablation Studies}
\paragraph{Discrete Token Conditioning.}  
In our approach, built on the MAR~\cite{li2025autoregressive} framework, we incorporate discrete tokens as conditions and modify the continuous tokenizer to VAVAE~\cite{yao2025reconstruction}. To isolate the effect of discrete conditioning, we also conduct experiments by replacing the VAVAE with the LDM~\cite{rombach2022high} tokenizer used in the original MAR. Table~\ref{tab:mar} presents this ablation study results on discrete token conditioning. In these experiments, we augment the MAR model with discrete tokens under various configurations, including parameter sizes and the number of AR steps, to investigate their impact on generation performance. The results reveal that incorporating discrete tokens reduces the gFID by up to 0.2 points and boosts the IS by approximately 10 points, indicating that discrete conditioning substantially improves overall generation quality. This improvement can be attributed to modeling the conditional probability from discrete to continuous tokens, which simplifies the optimization process and enables the generation of higher-quality images with better fine details. 

Furthermore, the results demonstrate that discrete token conditioning significantly reduces the number of AR steps required for high-quality generation. While the original MAR models require 256 steps, our approach achieves superior performance with as few as 16–32 steps, leading to a substantial speedup. As shown in Table~\ref{tab:mar}, the inference time per image is reduced by approximately 5× compared to the original MAR models, even when accounting for the additional discrete token generation step. This efficiency gain stems from the fact that discrete tokens provide strong structural priors, allowing the continuous AR model to converge with fewer steps. As a result, our method achieves both higher generation quality and faster inference.
\begin{table}[t]
    \centering
    \tabcolsep=0.15cm
    \begin{tabular}{cccccc}
    \toprule
    Method & Params & gFID$\downarrow$ & IS $\uparrow$ & Steps & sec/img \\
    \midrule
    MAR-B & 208M & 2.31 & 281.7 & 256 & 0.866 \\
    MAR-L & 479M & 1.78 & 296.0 & 256 & 1.211 \\
    MAR-H & 943M & 1.55 & 303.7 & 256 & 1.678 \\
    \midrule
    + Condition & 184M & 1.86 & 292.0 & 16 & 0.195 \\
    + Condition & 427M & 1.57 & 309.7 & 16 & 0.203 \\
    + Condition & 558M & \textbf{1.40} & 324.7 & 32 & 0.316 \\
    \bottomrule
    \end{tabular}
    \caption{Ablation study results on discrete token conditioning. We condition MAR on discrete tokens across various settings, including different parameter sizes and AR steps. Note that these results are preliminary; additional training time and AR steps are expected to further improve performance. The sec/image values were measured using a batch size of 100, and our method’s inference time accounts for the discrete token generation. \emph{Params} denotes the number of trainable parameters.}
    \label{tab:mar}
\end{table}
\begin{table}[t]
    \centering
    \tabcolsep=0.04cm
    \begin{tabular}{ccc|ccc}
    \toprule
    \makecell{Discrete \\ AR Model} & gFID$\downarrow$ & sec/img & \makecell{+ Continuous \\ AR Model} & gFID$\downarrow$ & $^\star$sec/img \\
    \midrule
    \multirow{2}{*}{*RAR-B} & \multirow{2}{*}{1.97} & \multirow{2}{*}{0.080} & DisCon-B & 1.91 & 0.149 \\
      &  &  & DisCon-L &  1.87 & 0.171  \\
    \midrule
    \multirow{2}{*}{*RAR-L} & \multirow{2}{*}{1.74} & \multirow{2}{*}{0.085} & DisCon-B & 1.74 & 0.143 \\
      &  &  & DisCon-L & 1.71  & 0.176  \\
    \midrule
    \multirow{2}{*}{*RAR-XXL} & \multirow{2}{*}{1.50} & \multirow{2}{*}{0.145} & DisCon-B & 1.41 & 0.203 \\
      &  &  & DisCon-L &  \textbf{1.38} &  0.236 \\
    \bottomrule
    \end{tabular}
    \caption{Ablation study results on discrete AR models used for generating conditioning tokens. *The RAR results are obtained on our device, which slightly differ from the original paper. $^\star$Our method’s inference time accounts for the discrete token generation. Note that the results under RAR-B and RAR-L show limited improvements, suggesting these models may not capture discrete representations well, leading to imperfect conditioning. Conversely, a stronger discrete AR model appears to further enhance the final generation performance of our method.}
    \label{tab:condition}
\end{table}
\begin{figure}
    \centering
    \includegraphics[width=\linewidth]{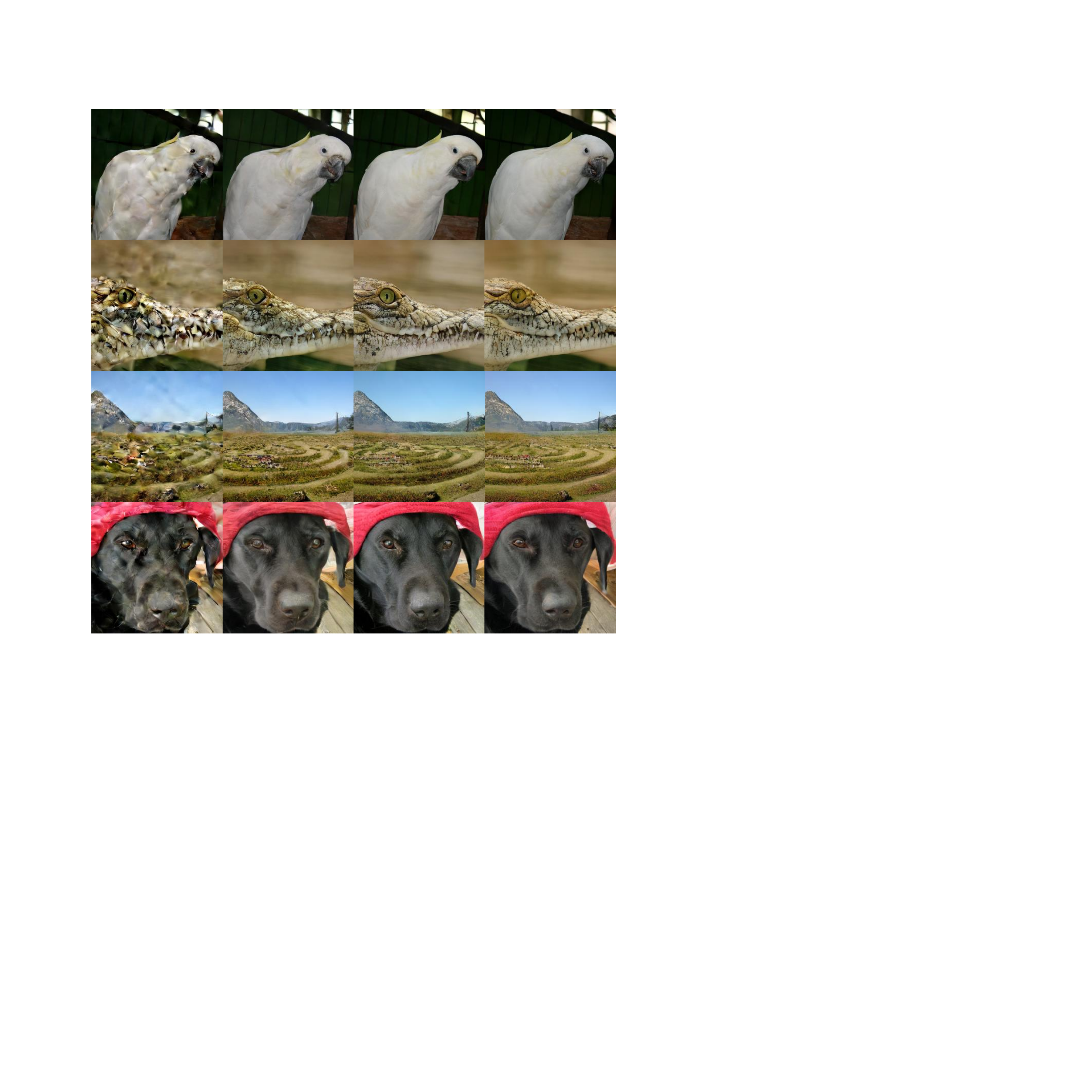}
    \captionof{figure}{\textbf{Results with Different AR Steps.} From left to right, images are generated using \{1, 4, 16, 32\} AR steps. Notably, plausible results can be generated with as few as 4 steps.}
    \label{fig:step_vis}
\end{figure}
\begin{figure*}[t]
  \begin{minipage}[t]{0.318\linewidth}
    \vspace*{0pt} 
    \centering
    \includegraphics[width=\linewidth]{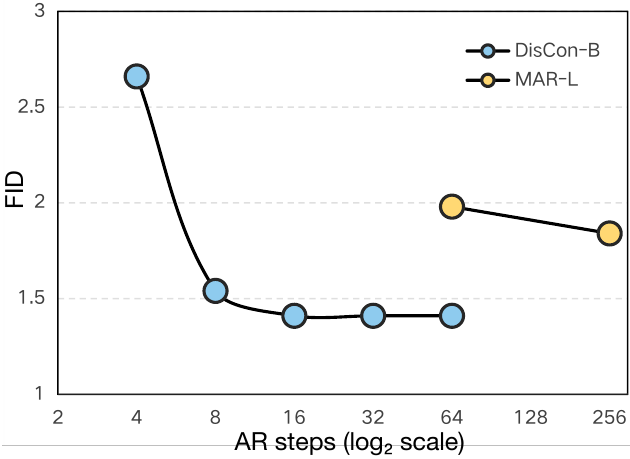}
    \captionof{figure}{\textbf{Effect of AR Steps on Generation Performance.} 
    Our method achieves stable performance with more than 16 steps, whereas MAR degrades when reducing steps from 256 to 64. The results for MAR-L are taken from its original paper. Note that both DisCon-B and MAR-L employ approximately 400M parameters.}
    \label{fig:step}
  \end{minipage}\hfill
  \begin{minipage}[t]{0.32\linewidth}
    \vspace*{0pt} 
    \centering
    \includegraphics[width=\linewidth]{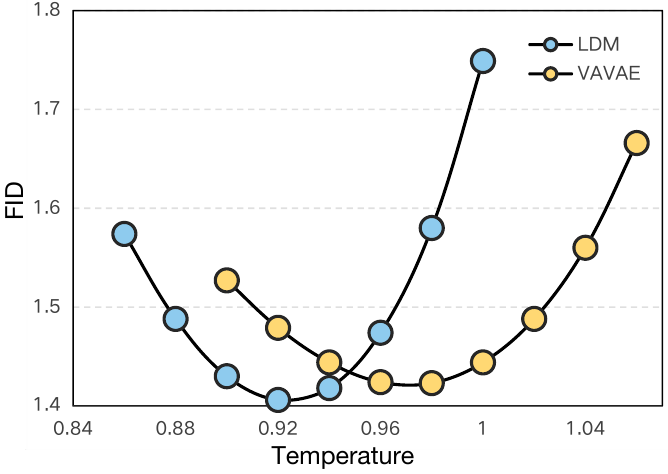}
    \captionof{figure}{\textbf{Temperature of the Diffusion Model.} Temperature critically affects generation quality: lower values yield more deterministic and high-fidelity outputs, while higher values increase diversity at the expense of image quality. The optimal setting varies with the continuous tokenizer used, as indicated by the legend. }
    \label{fig:temp}
  \end{minipage}\hfill
  \begin{minipage}[t]{0.322\linewidth}
    \vspace*{0pt} 
    \centering
    \includegraphics[width=\linewidth]{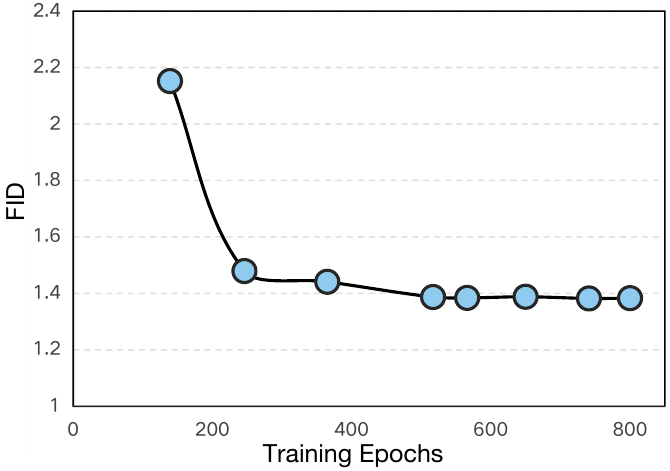}
    \captionof{figure}{\textbf{Generation Performance during Training.} Our method achieves SOTA performance (around 1.5 gFID) after 200 training epochs. Note that these results are obtained with a 0.9999 EMA setting, which may cause a slight performance lag. Detailed training loss curves in the supplementary materials provide additional insights.}
    \label{fig:epoch}
  \end{minipage}
\end{figure*}
\paragraph{Discrete AR Models.}  
Table~\ref{tab:condition} presents ablation study results on various discrete AR models used for generating conditioning tokens. As noted in the table caption, our experiments show that weaker models, RAR-B and RAR-L, offer only marginal improvements. This is likely due to their limited ability to model discrete representations, which results in imperfect conditioning for the continuous AR model. In contrast, a stronger model like RAR-XXL produces more precise discrete tokens, which in turn leads to noticeably better generation performance in DisCon. These results highlight the importance of high-quality discrete tokens, as they serve as reliable structural priors that enhance both consistency and detail synthesis. Thus, investing in more powerful discrete AR models can directly enhance the overall performance of our method. Note that our method exhibits a slightly slower inference speed, since it involves both discrete and continuous token generation. However, thanks to the efficiency of our continuous token generation process, which requires only a few AR steps, the overall inference speed remains competitive with purely discrete models like RAR.
\paragraph{Autoregressive Steps.}  
Previous methods reveal that increasing the number of AR steps generally improves generation quality; however, the incorporation of discrete tokens enables our method to maintain robust performance with far fewer steps. For instance, while MAR-H requires 256 AR steps to achieve a gFID of 1.55, our approach achieves a better gFID 1.38 using only 16 steps (the discrete AR steps are excluded from the comparison since their inference time is negligible). As illustrated in Figure~\ref{fig:step}, our method reaches stable performance at 16 steps, whereas MAR's performance degrades when its AR steps are reduced from 256 to 64. Furthermore, the qualitative results shown in Figure~\ref{fig:step_vis} indicate that even with as few as 4 AR steps, our method produces plausible outputs with reasonable global consistency. 
These observations underscore that the discrete tokens provide strong structural guidance, effectively reducing the reliance on numerous autoregressive iterations and thereby lowering computational cost without sacrificing image quality.
\paragraph{Diffusion Model Settings.}  
The configurations of classifier-free guidance (CFG) and temperature are critical for diffusion model performance. In our implementation, CFG is applied during the discrete token generation phase, while applying CFG to continuous token generation leads to degraded performance. Moreover, our experiments indicate that the optimal temperature setting is dependent on the continuous tokenizer used, as illustrated in Figure~\ref{fig:temp}. 
\paragraph{Training Epochs.}
Figure~\ref{fig:epoch} shows the generation performance of our method during training. Our model achieves SOTA performance (around 1.5 gFID) after only 200 training epochs, although the use of a 0.9999 EMA (Exponential Moving Average) may introduce a performance lag. This training efficiency demonstrates the reduced optimization complexity achieved by modeling the conditional probability from the discrete distribution to the continuous distribution. Detailed training loss curves in the supplementary materials further support these findings.



\subsection{Discussion}
Our innovation lies in simplifying the modeling of continuous distributions into two distinct steps: first, modeling the discrete distribution, and second, modeling the conditional probability from discrete to continuous tokens. This two-step formulation reduces the optimization complexity and enables our model to generate high-quality images with improved fidelity and reconstruction accuracy. Experimental results show that incorporating discrete token conditioning reduces the gFID by up to 0.2 points and boosts the IS by approximately 10 points. For instance, our DisCon-L model achieves a gFID of 1.38, surpassing the state-of-the-art discrete AR model RAR-XXL, which achieves a gFID of 1.48. Although our method involves a two-stage generation process—first generating discrete tokens as conditional signals and then performing continuous AR with a reduced number of steps—the overall inference speed remains competitive. Moreover, as an autoregressive approach, our method is inherently compatible with large language models (LLMs) and can be seamlessly integrated into multimodal frameworks, offering a key advantage over diffusion models.

\section{Conclusion} In this paper, we introduced DisCon, a novel autoregressive framework that simplifies the modeling of continuous distributions by decoupling the task into two sequential steps: first, modeling the discrete distribution, and second, learning the conditional probability from discrete tokens to continuous tokens. This formulation alleviates the optimization challenges associated with continuous representations while avoiding the information loss induced by direct quantization. Quantitative evaluations on ImageNet-256 demonstrate that our approach outperforms state-of-the-art visual AR models in both generation fidelity and reconstruction quality. Moreover, as an autoregressive model, DisCon supports high compatibility with large language models, distinguishing it from diffusion-based approaches and paving the way for future multimodal applications. We believe that DisCon opens promising avenues for high-fidelity image synthesis, and further exploration of model scaling and integration with LLMs will unlock even greater potential.
\section*{Acknowledgments}

This work is supported in part by the Hong Kong Research Grants Council under the NSFC/RGC Collaborative Research Scheme (Grant CRS\_HKU703/24), the National Natural Science Foundation of China (No. 62202199), and the Fundamental Research Funds for the Central Universities.
{
    \small
    \bibliographystyle{ieeenat_fullname}
    \bibliography{main}
}

\newpage
\twocolumn[{%
    \centering
    {\LARGE \textbf{Supplementary Materials}\par}
    \vspace{1em} 
    \includegraphics[width=\linewidth]{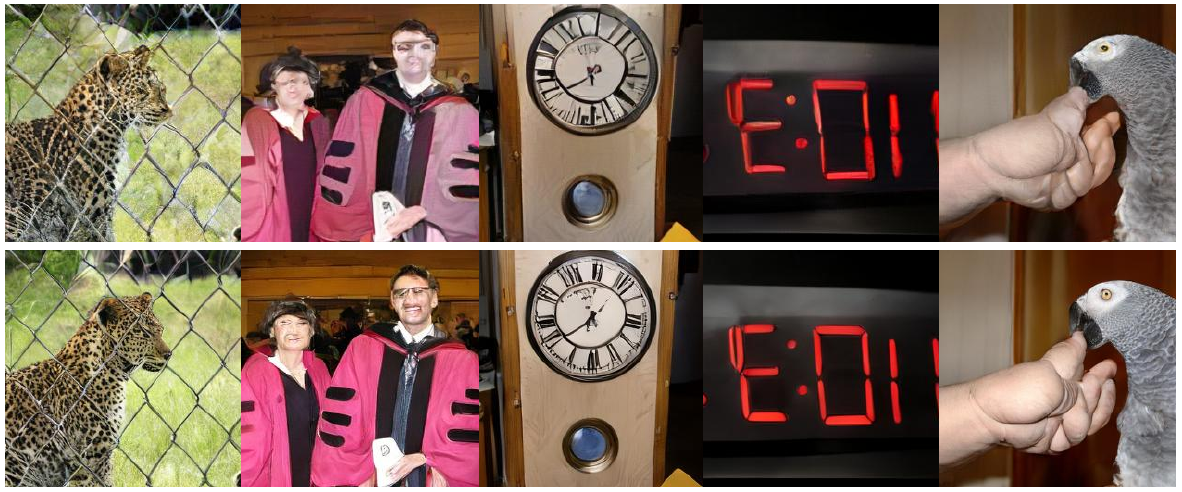}
    \captionof{figure}{\textbf{Demonstration of Failure Cases.} Top: Images generated by RAR-XXL. Bottom: Images generated by our proposed DisCon-L. Although such issues are common in image synthesis, our method exhibits improved performance.}
    \label{fig:failure}
    \vspace{1em} 
}]

\appendix

\section{Implementation Details}
In our experiments, the training is conducted for a default of 800 epochs. For DisCon-L, the batch size per GPU is set to 56, whereas for DisCon-B it is set to 90. Both models share a backbone with 32 transformer blocks and a width of 1024. The primary difference between the two lies in the diffusion head: DisCon-L employs 12 blocks with a width of 1536, while DisCon-B uses 3 blocks with a width of 1024.

\section{Failure Cases}
Figure~\ref{fig:failure} illustrates typical failure cases observed in both our method and RAR-XXL, including challenges with human faces, characters, and hands. It is important to emphasize that these issues are inherent to most image generation methods and are not unique to any single approach. Despite the prevalence of these common challenges, our approach consistently outperforms the SOTA RAR-XXL model.

\begin{figure}
    \centering
    \includegraphics[width=\linewidth]{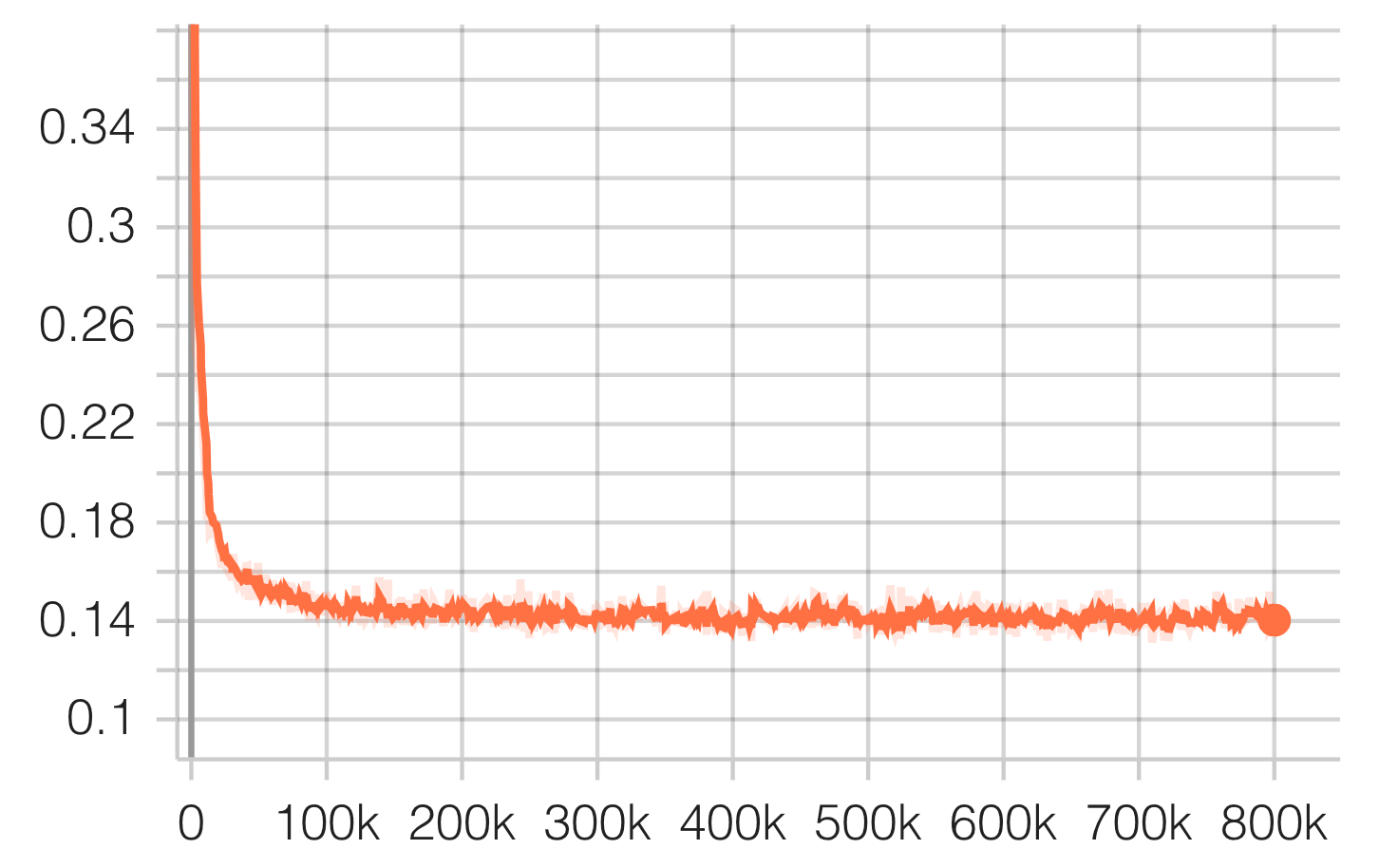}
    \caption{\textbf{Training Loss Curve of DisCon-L.} The loss converges at around 100 epochs, demonstrating the reduced optimization complexity of our approach.}
    \label{fig:loss}
\end{figure}

\section{Training Process}
Figure~\ref{fig:loss} shows the training loss curve for DisCon-L. The loss stabilizes at around 100 epochs, demonstrating the reduced optimization complexity achieved by our two-stage approach. 
The efficient training dynamics underscore the benefits of decoupling the modeling of discrete and continuous representations, leading to more reliable and high-quality image synthesis.

\section{Discrete AR Models}
Our method leverages discrete tokens generated by discrete AR models. We explore performance under different RAR models in Figure~\ref{fig:rar}. Additionally, we present inference results obtained by decoding these discrete tokens, which consistently demonstrate improved performance of our method.

\section{Generated Results}
\begin{figure*}
    \centering
    \includegraphics[width=\linewidth]{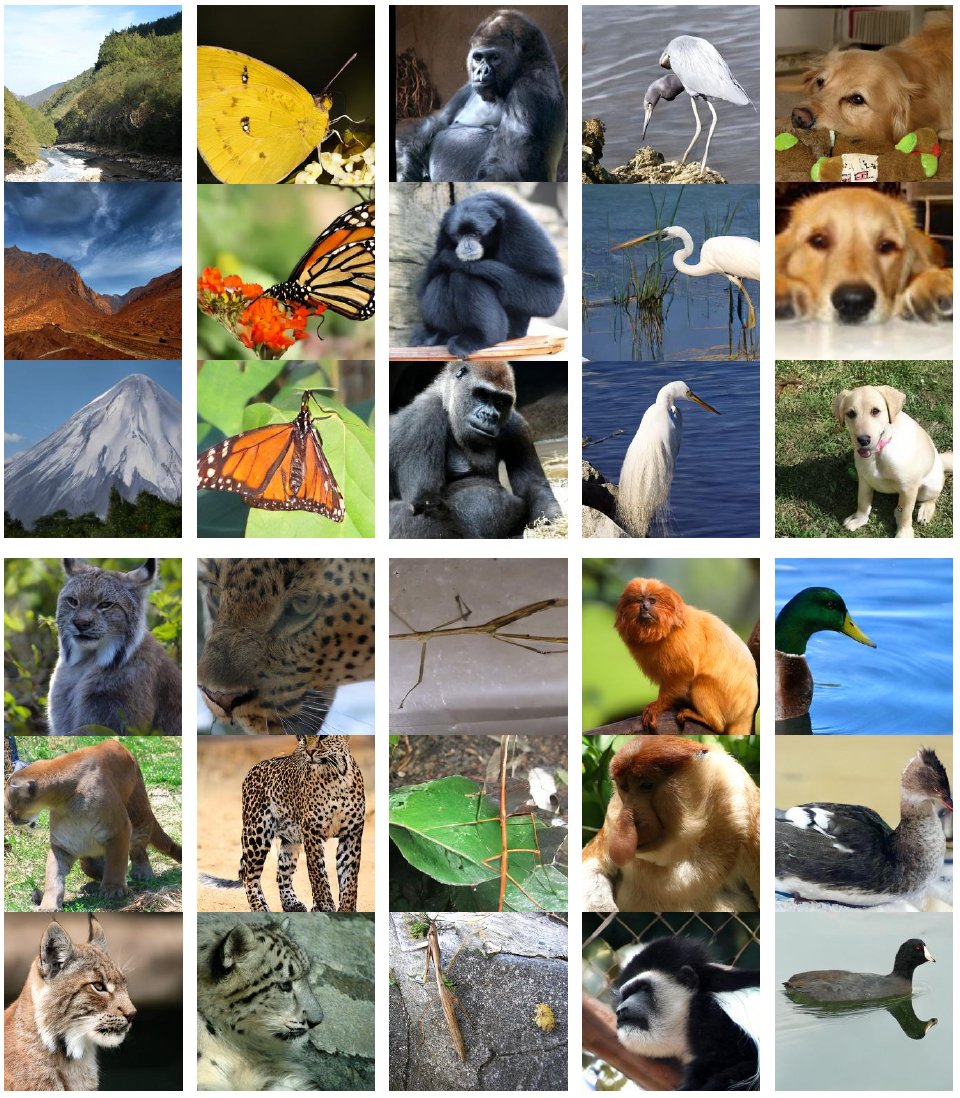}
    \caption{\textbf{Class-Conditioned Generation.} This figure showcases images generated by DisCon-L across various classes, demonstrating the high fidelity and diversity achieved by our approach.}
    \label{fig:discon}
\end{figure*}

Figure~\ref{fig:discon} presents sample images generated under different class labels, showcasing the high-fidelity synthesis and diversity achieved by our method. 

\section{Limitations and Future Directions.}
Despite these advantages, our method still relies on a diffusion head for generating continuous tokens. Although we use a lightweight diffusion head to mitigate computational overhead, its inclusion inevitably impacts overall efficiency. Moreover, while the two-step approach simplifies the continuous modeling process, it may not be the optimal solution for all scenarios. Future research could explore alternative strategies to further balance efficiency and quality, as well as investigate novel conditioning mechanisms for continuous token generation to enhance image synthesis performance.

\begin{figure*}
    \centering
    \includegraphics[width=\linewidth]{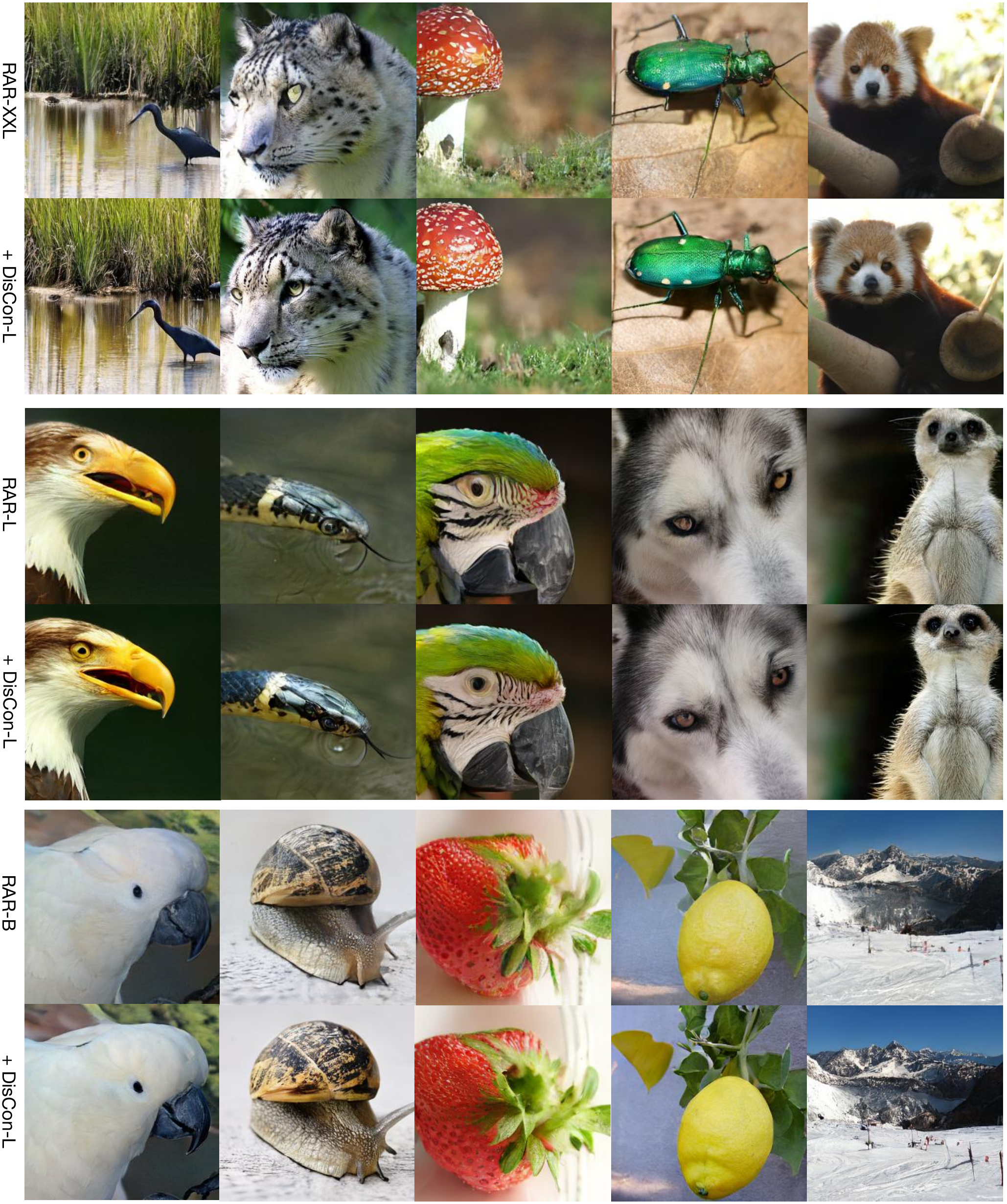}
    \caption{\textbf{Results conditioned on discrete tokens generated by different AR models.} From top to bottom: RAR-XXL, RAR-L, and RAR-B. For each model, the top row shows results generated by the respective RAR model, while the bottom row displays outputs from our DisCon method. Zoom in for better visualization to observe the significant improvements in generation quality.}
    \label{fig:rar}
\end{figure*}

\end{document}